\documentclass{article}
\usepackage{spconf,amsmath,graphicx}
\usepackage{booktabs}
\usepackage{multirow}
\usepackage{amssymb}
\usepackage{enumitem} 
\usepackage{stfloats}
\usepackage{placeins}
\usepackage{balance}
\usepackage{float} 

\title{TAE: Target-Aware Enhancer for Nighttime UAV Tracking}
\name{Yanyan Chen, Ruigang Fu, Yu Song, Ping Zhong}
\address{College of Electrical Science and Technology, National University of Defense Technology}

\begin{document}
\maketitle

\begin{abstract}
	Severe image degradation under low-light nighttime conditions constitutes a core bottleneck preventing all-day applications for UAV-based single object tracking. Existing image enhancement methods often struggle to distinguish between target and background regions, which can easily lead to amplified background noise or compromise target features. To overcome this limitation, we propose TAE, a target-aware low-light enhancement framework tailored for nighttime object tracking. Guided explicitly by weak supervisory signals from tracking bounding boxes, the framework performs region-aware enhancement to ensure operations focus on the target area. It further adopts an adaptive RGB multi-curve fusion mechanism to achieve refined modeling and adaptive adjustment across different regions. To facilitate research in this domain, we also contribute DarkSOT, a new benchmark for nighttime UAV tracking, comprising 268 sequences across 9 target categories. Experimental results on the DarkSOT and UAVDark135 demonstrate that TAE significantly improves tracking performance in low-light nighttime scenarios, exhibiting strong robustness and generalization. The DarkSOT dataset is available at https://github.com/Fu0511/DarkSOT-Dataset.
\end{abstract}


\section{Introduction}

Single object tracking based on unmanned aerial vehicle (UAV) platforms is a core technology for applications such as intelligent surveillance and autonomous inspection~\cite{ref1}. While current tracking methods perform well under normal lighting conditions, their effectiveness often degrades significantly in nighttime scenarios due to degraded image quality and difficulty in feature extraction~\cite{ref2}. This limitation restricts the potential for all-day operation of UAV systems.

To improve tracking performance in nighttime low-light conditions, mainstream approaches typically employ an image enhancer as a preprocessing module for the tracker. Existing methods fall into two main categories. The first includes general image enhancement techniques~\cite{ref3,ref4,ref5}, designed to boost overall brightness, yet they neglect tracking-specific needs, risking background noise amplification or weakened target features. The second type includes tracking-oriented enhancement methods~\cite{ref6,ref7,ref8}, which usually embed lightweight enhancement modules into the tracking pipeline. Among them, HighlightNet~\cite{ref9} is a representative work that introduces a learnable range mask to concentrate attention on potential target areas for selective enhancement. However, its range mask relies on implicit learning, lacking explicit target-location supervision and limiting illumination modeling to a single Gamma curve.

\begin{figure}[!t]
	\centering
	\includegraphics[width=3.4in]{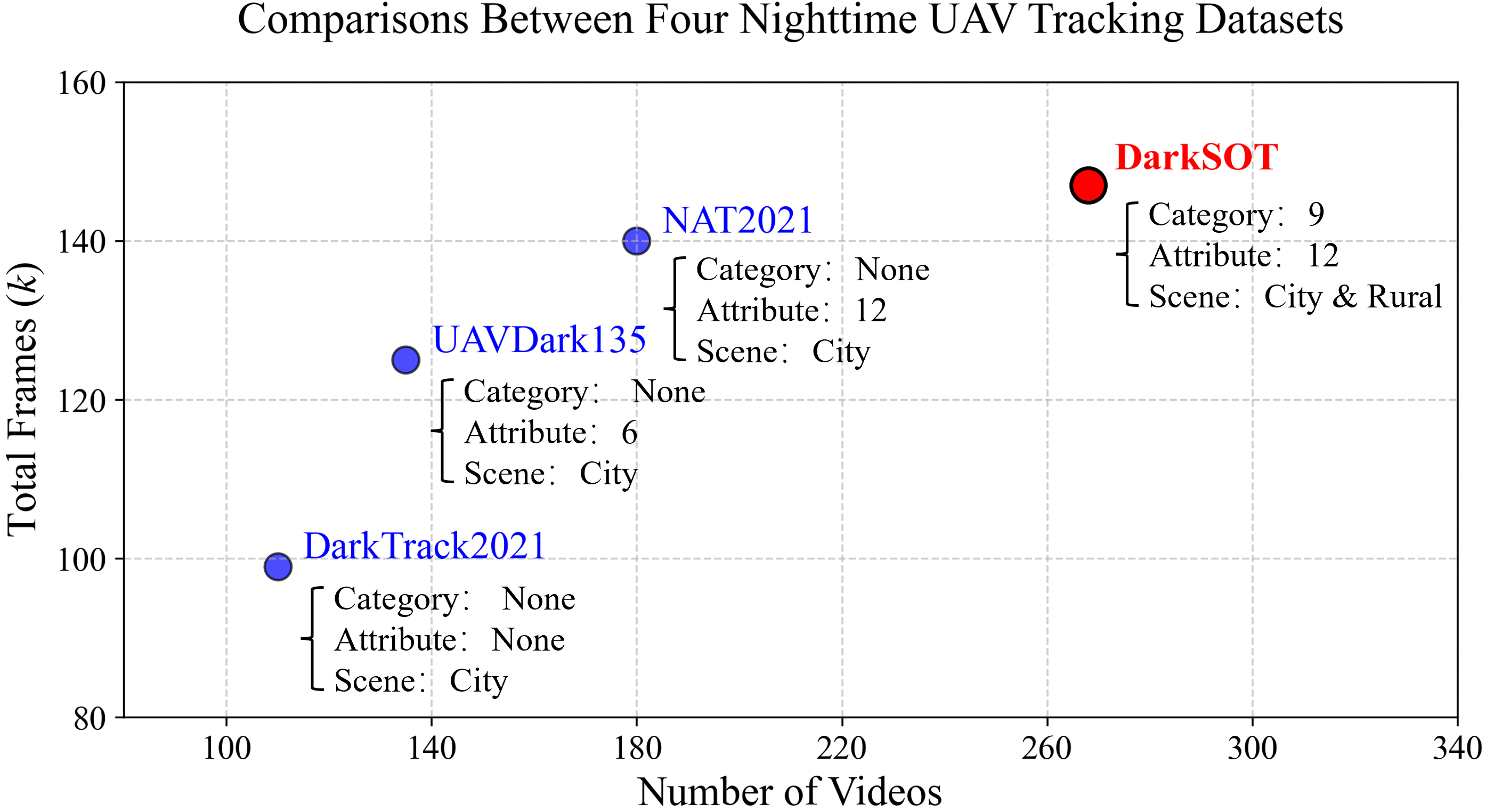}
	\caption{Overview of existing nighttime UAV tracking benchmarks with high-quality dense annotations, including  DarkTrack2021~\cite{ref7}, UAVDark135~\cite{ref10}, NAT2021~\cite{ref11}, and the proposed DarkSOT. DarkTrack2021 and UAVDark135 lack standardized object category definitions, while NAT2021 does not provide any object annotations. The proposed DarkSOT benchmark surpasses existing datasets in both scale and diversity. Best viewed in color.}
	\label{fig1}
\end{figure}

\begin{figure*}[!t]
	\centering
	\includegraphics[width=6.8in]{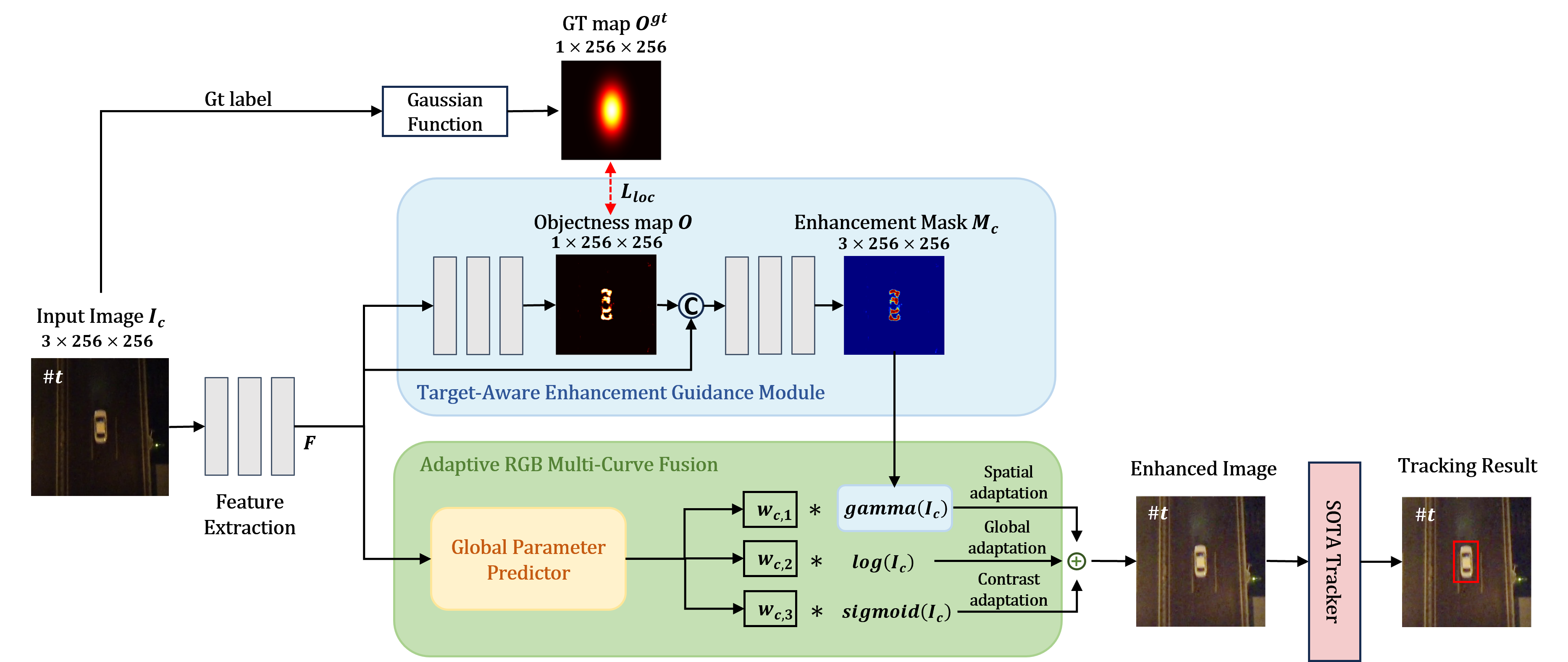}
	\caption{An overview of the TAE pipeline. The Target-Aware Enhancement Guidance Module generates an enhancement mask focused on the target region under the supervision of ground-truth boxes. The Adaptive RGB Multi-Curve Fusion module adaptively enhances and blends the image using three complementary curves.}
	\label{fig2}
\end{figure*}

To address the aforementioned limitations, we proposes TAE, a target-aware low-light enhancement framework for nighttime UAV tracking. On one hand, we design a target-aware enhancement guidance module, which utilizes weak supervision from ground-truth bounding boxes to generate a objectness prediction map, thereby providing explicit spatial guidance for the enhancement process. On the other hand, we introduce an adaptive RGB multi-curve  fusion mechanism. By dynamically fusing three complementary enhancement curves, including Gamma, logarithmic, and sigmoid curves, it enables refined and adaptive adjustment of illumination across different regions.

Furthermore, to advance research in nighttime tracking, we construct a new benchmark named DarkSOT, as shown in Fig .1. It contains 268 high-quality video sequences across 9 target categories, with clearly defined training and test splits. Compared to existing public datasets, DarkSOT offers richer scene diversity, a larger scale of sequences, and presents a more diverse set of challenges, thereby providing a comprehensive and demanding benchmark for developing and assessing nighttime tracking algorithms.

The main contributions of this work can be summarized as follows:
\begin{itemize}[itemsep=2pt, parsep=0pt, topsep=5pt]
	\item We propose TAE, the first enhancement framework to explicitly associate nighttime low-light enhancement with the tracking task via location supervision, ensuring enhancement concentrates on the target region.
	\item We design an adaptive RGB multi-curve fusion strategy with Gamma for target brightness, logarithmic for dark-region detail recovery, and sigmoid for contrast enhancement, providing stronger robustness than single-curve methods.
	\item We construct DarkSOT, the largest and most diverse nighttime UAV tracking benchmark, with 268 sequences, 9 categories, and 12 challenge attributes.
\end{itemize}

\section{Method}

This work proposes a target‑aware enhancer TAE to enhance potential target features at nighttime. As shown in Fig. 2, TAE contains two key modules: a Target‑Aware Enhancement Guidance Module that distinguishes target from background, and an Adaptive RGB Multi‑Curve Fusion Module that enables pixel‑wise enhancement. 
 
\subsection{Target-Aware Enhancement Guidance Module}

To precisely enhance targets and suppress background interference, we design a target-aware guidance module that focuses network attention on foreground objects.

\textbf{Objectness Probability Prediction:} 
The input image is denoted as $\mathbf{I}_c$, where $c \in \{R, G, B\}$. It passes through a feature extraction network to obtain the feature map $\mathbf{F}$. This map is then fed into an objectness prediction network to produce the pixel-wise objectness map $\mathbf{O} \in [0,1]^{1 \times H \times W}$, which indicates the likelihood of each pixel belonging to the foreground target. It is trained under the weak supervision of the ground-truth bounding box $\mathbf{b} = (x, y, w, h)$ to guide the network's attention toward the target region.

To avoid gradient instability caused by hard labels, we convert the ground-truth box into a Gaussian-based soft label $\mathbf{O}^{gt}$. The box center is computed as:
\begin{equation}
	(c_x, c_y) = (x + w/2, y + h/2)
\end{equation}

The soft label is then generated by:
\begin{equation}
	\mathbf{O}^{gt}(i,j) = \exp\left(-\frac{1}{2}\left[\left(\frac{i - c_x}{\sigma_x}\right)^2 + \left(\frac{j - c_y}{\sigma_y}\right)^2\right]\right)
	\label{eq:soft_label}
\end{equation}
where $\sigma_x = w/2$ and $\sigma_y = h/2$. By aligning the Gaussian distribution with the target size, this design concentrates enhancement attention toward the target center while maintaining smooth transitions at the boundaries to facilitate stable training.

The objectness prediction is optimized by a composite loss function $\mathcal{L}_{\text{loc}}$:
\begin{equation}
	\begin{split}
		\mathcal{L}_{\text{loc}} = &-\sum_{i,j}\left[\mathbf{O}^{gt}\log\mathbf{O} + (1-\mathbf{O}^{gt})\log(1-\mathbf{O})\right] \\
		&+ 1 - \frac{2\sum\mathbf{O}\mathbf{O}^{gt}}{\sum\mathbf{O} + \sum\mathbf{O}^{gt}}
	\end{split}
\end{equation}
The first term is a cross-entropy loss that provides pixel-wise supervision, while the second term is a Dice loss that emphasizes regional overlap. These complementary terms jointly optimize both boundary precision and regional consistency of the objectness map.

\textbf{Adaptive Enhancement Mask Generation:}
Based on the objectness map $\mathbf{O}$, a lightweight CNN generates the spatially-adaptive enhancement mask $\mathbf{M}_c \in \mathbb{R}^{ 3\times H \times W}$. This mask assigns higher enhancement weights to target regions while moderately adjusting background regions to suppress noise amplification. The mask is computed as:
\begin{equation}
	\mathbf{M}_c = f_{\text{CNN}}(\text{Concate}(\mathbf{O}, \mathbf{F}))
\end{equation}
where $f_{\text{CNN}}$ is a 3-layer convolutional network and $\mathbf{F}$ is the original feature from the backbone. This design enables target-aware enhancement: foreground objects receive aggressive enhancement to maximize visibility, while background regions are processed conservatively to preserve overall image quality and prevent noise amplification.

\subsection{Adaptive RGB Multi-Curve Fusion}

Low-light scenes exhibit diverse degradation patterns across spatial regions and color channels. The baseline method HighlightNet~\cite{ref9} employs a single Gamma curve for global enhancement, which is insufficient to handle such variability, particularly in complex low-light tracking scenarios. To address this limitation, we propose an adaptive multi-curve fusion strategy that performs collaborative optimization through three complementary curves across three dimensions: spatial adaptation, global adaptation, and contrast adaptation.

For each color channel, we define three complementary curve transformations:

\textbf{Gamma curve} for spatial-adaptive brightness enhancement:
\begin{equation}
	\begin{split}
		&\operatorname{Gamma}(\mathbf{I}_c) = \mathcal{C}_1(\mathbf{I}_c) = (\mathbf{I}_c + \epsilon)^{\gamma_c}, \\
		&\gamma_c = \alpha_c^{\mathbf{M}_c}
	\end{split}
	\label{eq:gamma_curve}
\end{equation}
where $\epsilon = 10^{-3}$ ensures numerical stability. The power-law form models the nonlinear response of camera sensors~\cite{ref12}, 
with $\gamma_c < 1$ proportionally amplifying dark regions while mildly adjusting bright regions. The learnable parameter $\alpha_c $
is spatially modulated by the normalized target-aware enhancement mask $\mathbf{M}_c \in [0,1]$ 
from Eq.~(4), yielding $\gamma_c$ constrained between $\alpha_c$ and 1. This enables aggressive brightening on targets while preserving 
background naturalness.

\textbf{Logarithmic curve} for  global brightness enhancement:
\begin{equation}
	Log(\mathbf{I}_c) = \mathcal{C}_2(\mathbf{I}_c) = \frac{\log(1 + 10\mathbf{I}_c)}{\log(11)}
\end{equation}
Based on Retinex theory~\cite{ref13}, the logarithmic transformation employs a large slope in dark regions to increase the signal-to-noise ratio, effectively recovering low-light details critical for tracking.

\textbf{Sigmoidal curve} for contrast enhancement:
\begin{equation}
	Sigmoid(\mathbf{I}_c) = \mathcal{C}_3(\mathbf{I}_c) = \frac{1}{1 + \exp(-10(\mathbf{I}_c - 0.5))}
\end{equation}
This sigmoid transform extends the dynamic range in the mid-intensity range to enhance local contrast, with the flat characteristics at both ends ensuring the stability of extreme brightness areas and preventing excessive noise amplification~\cite{ref14}.

The final enhanced frame is computed as:
\begin{equation}
	\mathbf{I}^{enh}_c = \sum_{k=1}^{3} w_{c,k} \cdot \mathcal{C}_k(\mathbf{I}_c), \quad w_{c,k} = \frac{\exp(\hat{w}_{c,k})}{\sum_{k'}\exp(\hat{w}_{c,k'})}
\end{equation}
where fusion weights  $\{\hat{w}_{c,k}\}$ are predicted by the Global Parameter Predictor—a lightweight CNN that analyzes global temporal features. The final normalized weights  $\{{w}_{c,k}\}$ are then obtained via the softmax operation. This adaptive fusion enables the network to leverage the complementary properties of the three curves for robust enhancement across diverse low-light scenarios.

\subsection{Loss Functions}

Four loss functions are employed to jointly train the framework. Specifically, three loss functions are adopted from Zero\-DCE++~\cite{ref15}: the exposure control loss $L_{\text{exp}}$, the color consistency loss $L_{\text{color}}$, and the illumination smoothness loss $L_{\text{tv}}$. These are combined with the proposed target-aware localization loss $L_{\text{loc}}$ to form the complete training objective:
\begin{equation}
	L = \lambda_{loc} L_{\text{loc}} + \lambda_1 L_{\text{exp}} + \lambda_2 L_{\text{color}} + \lambda_3 L_{\text{tv}},
	\label{eq:total_loss}
\end{equation}
where $\lambda_{loc}$, $\lambda_1$, $\lambda_2$, and $\lambda_3$ are weighting coefficients that balance the contribution of each term.

\section{The DarkSOT Benchmark}
To advance nighttime UAV tracking research, we present DarkSOT, a dedicated benchmark dataset consisting of 268 annotated video sequences for low-light algorithm development and evaluation. Each sequence is annotated with 12 challenging attributes. The dataset includes nine common target categories: People, Tricycle, Bicycle, Bus, Truck, Awning-tricycle, Motor, Car, and Van. All data were captured using a DJI M30T UAV platform under varied real-world nighttime conditions, with flight altitudes ranging from 10 to 100 meters and camera pitch angles between 15 and 90 degrees. This ensures rich diversity in viewpoint and scene content. The dataset is divided into 192 training sequences and 76 testing sequences. It provides a comprehensive and reproducible benchmark for the community. Compared to existing datasets, DarkSOT is larger in scale, more diverse in scenes, and presents greater low-light challenges, thereby facilitating the development of more robust tracking methods. Further details are provided in the Appendix.

\begin{figure*}[!t]
	\centering
	\includegraphics[width=6.5in]{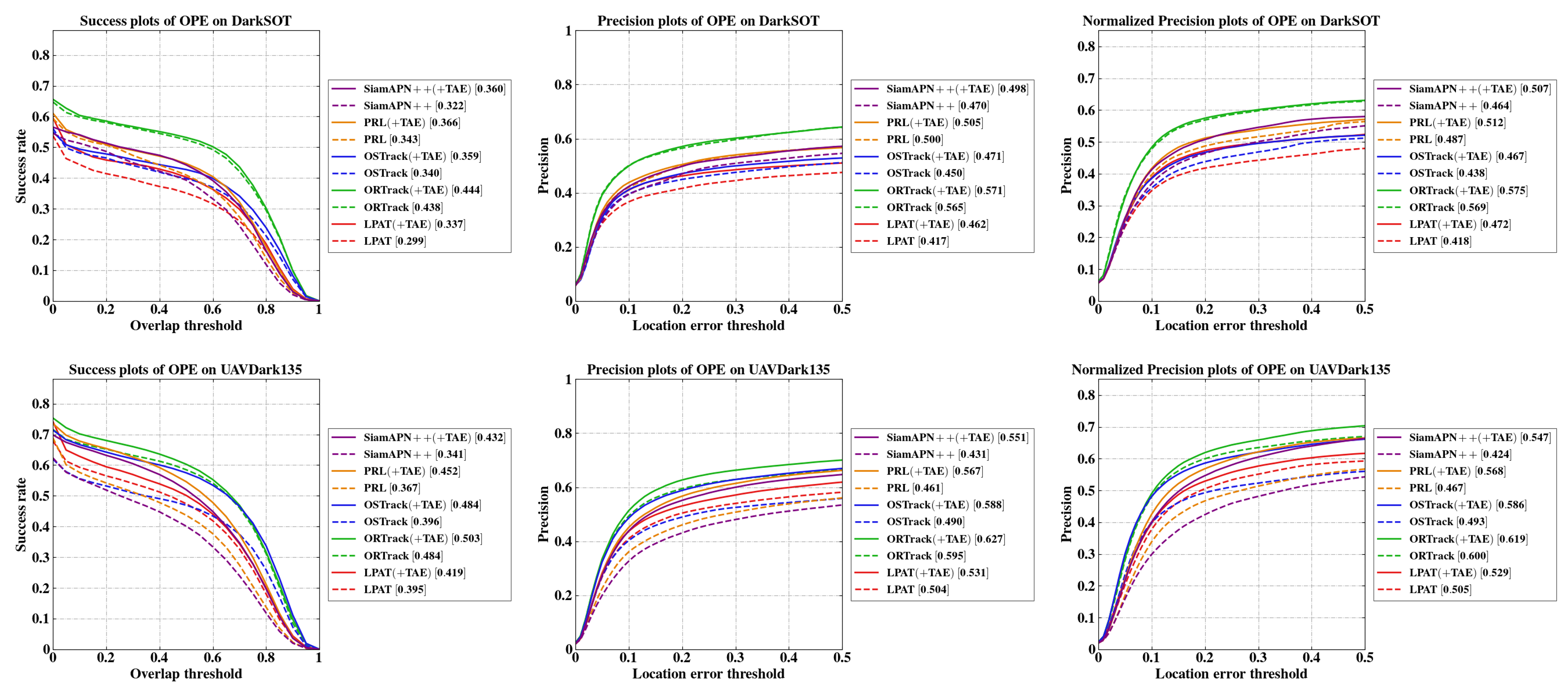}
	\caption{Overall performance of SOTA trackers with or without TAE on DarkSOT and UAVDark135~\cite{ref10} benchmarks.}
	\label{fig3}
\end{figure*}

\begin{table*}[!t] 
	\centering 
	\vspace{-15pt}
	\caption{Quantitative comparison results of SOTA trackers with and without TAE on two nighttime UAV datasets. } 
	\label{tab:algorithm_performance} 
	\resizebox{0.95\textwidth}{!}{
		\begin{tabular}{@{} l cccccccccccc @{}} 
			\toprule 
			\multirow{2}{*}{\textbf{Trackers}} & \multicolumn{6}{c}{\textbf{DarkSOT}} & \multicolumn{6}{c}{\textbf{UAVDark135}} \\ 
			\cmidrule(lr){2-7} \cmidrule(lr){8-13} 
			& \textbf{$S_{\text{AUC}}$} & \textbf{$\Delta$(\%)} & \textbf{P} & \textbf{$\Delta$(\%)} & \textbf{NormP.} & \textbf{$\Delta$(\%)} & \textbf{$S_{\text{AUC}}$} & \textbf{$\Delta$(\%)} & \textbf{P} & \textbf{$\Delta$(\%)} & \textbf{NormP.} & \textbf{$\Delta$(\%)}\\ 
			\midrule 
			LAPT~\cite{ref16} & 0.299 & - & 0.417 & - & 0.418 & - & 0.395 & - & 0.504 & - & 0.505 & -  \\ 
			LAPT(+TAE) & 0.337 & \textbf{3.8} & 0.462 & \textbf{4.5} & 0.472 & \textbf{5.4} & 0.419 & \textbf{2.4} & 0.531 & \textbf{2.7} & 0.529 & \textbf{2.4}  \\
			SiamAPN++~\cite{ref17} & 0.322 & - & 0.470 & - & 0.464 & - & 0.341 & - & 0.431 & - & 0.424 & - \\ 
			SiamAPN++(+TAE) & 0.360 & \textbf{3.8} & 0.498 & \textbf{2.8} & 0.507 & \textbf{4.3} & 0.432 & \textbf{9.1} & 0.551 & \textbf{12.0} & 0.547 & \textbf{12.3} \\
			PRL~\cite{ref18} & 0.343 & - & 0.500 & - & 0.487 & - & 0.367 & - & 0.461 & - & 0.467 & - \\ 
			PRL(+TAE) & 0.366 & \textbf{2.3} & 0.505 & \textbf{0.5} & 0.512 & \textbf{2.5} & 0.452 & \textbf{8.5} & 0.567 & \textbf{10.6} & 0.568 & \textbf{10.1}  \\ 
			OSTrack~\cite{ref19} & 0.340 & - & 0.450 & - & 0.438 & - & 0.396 & - & 0.490 & - & 0.493 & -  \\ 
			OSTrack(+TAE) & 0.359 & \textbf{1.9} & 0.471 & \textbf{2.1} & 0.467 & \textbf{2.9} & 0.484 & \textbf{8.8} & 0.588 & \textbf{9.8} & 0.586 & \textbf{9.3}  \\ 
			ORTrack~\cite{ref20} & 0.438 & - & 0.565 & - & 0.569 & - & 0.484 & - & 0.595 & - & 0.600 & -  \\ 
			ORTrack(+TAE) & 0.444 & \textbf{0.6} & 0.571 & \textbf{0.6} & 0.575 & \textbf{0.6} & 0.503 & \textbf{1.9} & 0.627 & \textbf{3.2} & 0.619 & \textbf{1.9}  \\  
			\bottomrule 
		\end{tabular} 
	}
\end{table*}

\section{Experiments}

\subsection{Implement Details}
The proposed TAE was trained on an NVIDIA A100 GPU using the PyTorch framework for 30 epochs. It was trained on the training split of our DarkSOT dataset. During training, input images were resized to 256×256 resolution. The model was optimized using the AdamW optimizer with an initial learning rate of 0.0001 and a weight decay of 0.0001. The batch size was set to 16. The weighting coefficients $\lambda_{loc}$, $\lambda_1$, $\lambda_2$, and $\lambda_3$ are set to 1.0, 1.0, 0.2, 0.1, respectively.

\subsection{Evaluation Metric}
Following the standard OTB protocol~\cite{ref21}, we perform a one-pass evaluation (OPE), where the tracker is initialized with the ground-truth in the first frame and runs without template updates thereafter. Three widely-used metrics are employed: AUC, Precision, and Normalized Precision. AUC measures the overall success rate across different IoU thresholds. Precision calculates the center location error between predicted and ground-truth boxes, while Normalized Precision normalizes this error by target size for scale-invariant accuracy.

\subsection{TAE in SOTA Trackers}
To evaluate the generalization of TAE across different trackers, we conduct experiments using five state-of-the-art trackers, including LAPT~\cite{ref16}, SiamAPN++~\cite{ref17}, PRL~\cite{ref18}, OSTrack~\cite{ref19} and ORTrack~\cite{ref20}. The overall performance of these trackers with and without TAE is presented in Table 1 and Fig. 3, based on evaluations on UAVDark135~\cite{ref10} and our newly introduced DarkSOT benchmark. The results indicate that TAE consistently improves nighttime tracking performance across all tested trackers on both benchmarks. For example, on UAVDark135, SiamAPN++ achieved gains of 9.1\% in success rate and 12.0\% in precision. On DarkSOT, LAPT improved by 3.8\% in success rate and 4.5\% in precision. Fig. 4 visually presents three selected sequences, further confirming TAE's consistent improvements in tracker performance under nighttime conditions.

\begin{figure}[!t]
	\centering
	\includegraphics[width=3.4in]{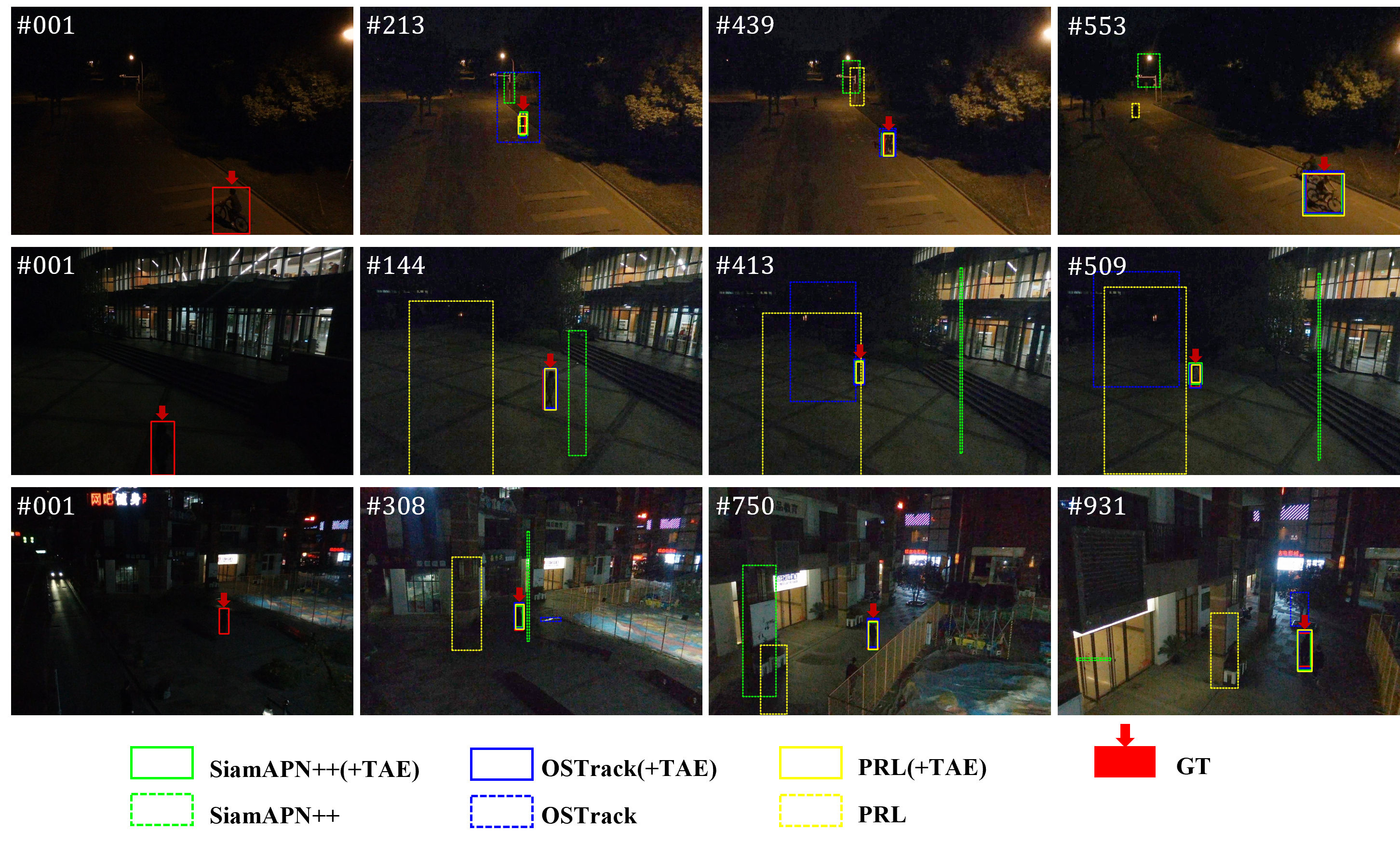}
	\caption{Qualitative tracking results are shown with solid boxes for trackers using TAE and dashed boxes for those without.}
	\label{fig4}
\end{figure}

\subsection{Comparison with SOTA Low-Light Enhancers}
To demonstrate the effectiveness of TAE in nighttime tracking, we compared it with nine state-of-the-art low-light enhancement methods: RUAS~\cite{ref3}, PairLIE~\cite{ref4}, SCI~\cite{ref5}, LDEnhancer~\cite{ref6}, SCT~\cite{ref7}, DarkLighter~\cite{ref8}, HighlightNet~\cite{ref9}, HVI-CIDNet~\cite{ref22}, and DarkIR~\cite{ref23}. Table~\ref{tab:enhancer} compares the tracking performance of these enhancers when integrated with the LAPT~\cite{ref16} tracker on the DarkSOT benchmark. The results show that TAE achieves the most significant improvement, outperforming all compared enhancers with gains of 3.8\% in success rate and 5.4\% in normalized precision.

\begin{table}[!t]
	\centering
	\caption{Performance comparison of different enhancement methods on DarkSOT dataset.}
	\label{tab:enhancer}
	\resizebox{0.46\textwidth}{!}{
		\begin{tabular}{@{} l c c c c @{}}
			\toprule
			\textbf{Method} & $S_{\text{AUC}}$ & \textbf{$\Delta_s$ (\%)} & NormP. & \textbf{$\Delta_n$ (\%)} \\
			\midrule
			LAPT & 0.299 & - & 0.418 & - \\
			\midrule
			+RUAS~\cite{ref3}         & 0.335 & 3.6 & 0.450 & 3.2 \\
			+PairLIE~\cite{ref4}      & 0.333 & 3.4 & 0.452 & 3.4 \\
			+SCI~\cite{ref5}          & 0.327 & 2.8 & 0.437 & 1.9 \\
			+LDEnhancer~\cite{ref6}   & 0.329 & 3.0 & 0.460 & 4.2 \\
			+SCT~\cite{ref7}          & 0.333 & 3.4 & 0.453 & 3.5 \\
			+DarkLighter~\cite{ref8}  & 0.330 & 3.1 & 0.449 & 3.1 \\
			+HighlightNet~\cite{ref9} & 0.307 & 0.8 & 0.423 & 0.5 \\
			+HVI-CIDNet~\cite{ref22}  & 0.324 & 2.5 & 0.437 & 1.9 \\
			+DarkIR~\cite{ref23}      & 0.309 & 1.0 & 0.424 & 0.6 \\
			\midrule
			+Ours & \textbf{0.337} & \textbf{3.8} & \textbf{0.472} & \textbf{5.4} \\
			\bottomrule
		\end{tabular}
	}
\end{table}

\subsection{Ablation Study}

This section presents an ablation study of the core components in TAE on the DarkSOT dataset using the LAPT tracker~\cite{ref16}. As summarized in Table 3, adding the target-aware (TA) guidance module yields success and norm precision 
rates of 0.318 and 0.443. Incorporating the multi-curve (MC) fusion module further improves performance by 3.8\% in success rate and 5.4\% in norm precision over the original LAPT tracker, confirming the contribution of each module.

We further analyze the sensitivity of the localization loss coefficient $\lambda_{loc}$, which controls the proposed $\mathcal{L}_{loc}$. As shown in Table 4, performance improves steadily as $\lambda_{loc}$ increases from 0 to 1.0, and degrades beyond that, confirming that $\lambda_{loc}=1.0$ is the optimal setting.

\begin{table}[!t] 
\centering
\caption{Ablation Study of the TAE on DarkSOT.}
\resizebox{0.46\textwidth}{!}{
\begin{tabular}{@{}lcccc@{}} 
	\toprule
	Trackers & $S_{\text{AUC}}$ & $\Delta_{s}(\%)$ & NormP. & \textbf{$\Delta_n$ (\%)} \\
	\midrule
	LAPT & 0.299 & -- & 0.418 & -- \\
	+Baseline & 0.307 & 0.8 & 0.423 & 0.5 \\
	+Baseline+TA & 0.318 & 1.9 & 0.443 & 2.5 \\
	\midrule
	\textbf{+Baseline+TA+MC} & \textbf{0.337} & \textbf{3.8} & \textbf{0.472} & \textbf{5.4} \\
	\bottomrule
\end{tabular}
}
\end{table}

\begin{table}[!t]
	\centering
	\caption{Sensitivity analysis of $\lambda_{loc}$ on DarkSOT.}
	\begin{tabular*}{0.8\columnwidth}{@{\extracolsep{\fill}}cccc@{}}
		\toprule
		$\lambda_{loc}$ & $S_{\text{AUC}}$ & P & NormP. \\
		\midrule
		0.0 & 0.307 & 0.417 & 0.418 \\
		0.5 & 0.320 & 0.440 & 0.447 \\
		\textbf{1.0} & \textbf{0.337} & \textbf{0.462} & \textbf{0.472} \\
		1.5 & 0.332 & 0.456 & 0.466 \\
		2.0 & 0.324 & 0.443 & 0.453 \\
		\bottomrule
	\end{tabular*}
\end{table}

\section{Conclusion}

This work proposes TAE, a target‑aware enhancement framework for nighttime UAV tracking. It employs tracking bounding boxes as weak supervision to focus enhancement on target regions, suppressing background noise while preserving discriminative features. An adaptive multi‑curve fusion mechanism further enables pixel-wise precise illumination adjustment. We also contribute DarkSOT, a dedicated nighttime UAV tracking benchmark. Compared to previous efforts, DarkSOT is currently the largest and most diverse UAV nighttime tracking dataset in terms of scale and scene variety. Comprehensive experiments on DarkSOT and UAVDark135 demonstrate that TAE significantly improves tracking performance with strong robustness and generalization, thereby advancing toward reliable all‑day UAV tracking capability.

\balance
\bibliographystyle{IEEEtran}
\bibliography{refs}

\end{document}